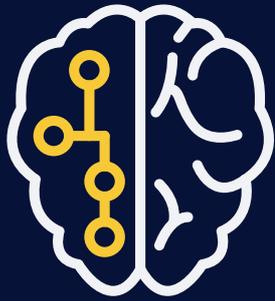

# Hybrid Intelligence

**Powered by Human Intuition.
Augmented by AI.**

CATCHWORD

# Hybrid Intelligence

Dominik Dellermann M.Sc. · Philipp Ebel · Matthias Söllner · Jan Marco Leimeister

## 1 Introduction

Research has a long history of discussing what is superior in predicting certain outcomes: statistical methods or the human brain. This debate has repeatedly been sparked off by the remarkable technological advances in the field of artificial intelligence (AI), such as solving tasks like object and speech recognition, achieving significant improvements in accuracy through deep-learning algorithms (Goodfellow et al. 2016), or combining various methods of computational intelligence, such as fuzzy logic, genetic algorithms, and case-based reasoning (Medsker 2012). One of the implicit promises that underlie these advancements is that machines will 1 day be capable of performing complex tasks or may even supersede humans in performing these tasks. This triggers new heated debates of when machines will ultimately replace humans (McAfee and Brynjolfsson 2017). While previous research has proved that AI performs well in some clearly defined tasks such as playing chess, playing Go or identifying objects on images, it is doubted that the development of an artificial general intelligence (AGI) which is able to solve multiple tasks at the same time can be achieved in the near future (e.g., Russell and Norvig 2016). Moreover, the use of AI to solve complex business problems in organizational contexts occurs scarcely, and applications for AI that solve complex problems remain mainly in laboratory settings instead of being implemented in practice.

Since the road to AGI is still a long one, we argue that the most likely paradigm for the division of labor between humans and machines in the next decades is Hybrid Intelligence. This concept aims at using the complementary strengths of human intelligence and AI, so that they can perform better than each of the two could separately (e.g., Kamar 2016).

## 2 Conceptual Foundations and What Hybrid Intelligence is Not

Before focusing on Hybrid Intelligence in detail, we first want to delineate the differences between this concept and related but still different forms of intelligence in this context.

### 2.1 Intelligence

Various definitions and dimensions (e.g., social, logical, spatial, musical) of the term intelligence exist in multiple research disciplines, such as psychology, cognitive science,

Accepted after two revisions by Prof. Weinhardt.

D. Dellermann M.Sc. · Prof. Dr. J. M. Leimeister
Research Center for IS Design (ITeG), Information Systems, University of Kassel, Pfannkuchstraße 1, 34121 Kassel, Germany

Dr. P. Ebel · Prof. Dr. M. Söllner ·
Prof. Dr. J. M. Leimeister (✉)
Institute of Information Management, University of St. Gallen, Müller-Friedberg-Strasse 8, 9000 St. Gallen, Switzerland
e-mail: janmarco.leimeister@unisg.ch

Prof. Dr. M. Söllner
Information Systems and Systems Engineering, University of Kassel, Henschelstraße 4, 34127 Kassel, Germany

Springer

neuro science, human behavior, education, or computer science. For the purpose of our research, we use an inclusive and generic definition to describe general intelligence. It is the ability to accomplish complex goals, learn, reason, and adaptively perform effective actions within an environment. This can generally be subsumed with the capacity to both acquire and apply knowledge (Gottfredson 1997). While intelligence is most commonly used in the context of humans (and more recently of intelligent artificial agents), it also applies to intelligent, goal-directed behavior of animals.

## 2.2 Human Intelligence

The sub-dimension of intelligence that is related to the human species defines the mental capabilities of human beings. On the most holistic level, it covers the capacity to learn, reason, and adaptively perform effective actions within an environment, based on existing knowledge. This allows humans to adapt to changing environments and act towards achieving their goals.

While one assumption concerning intelligence is the existence of a so-called "g-factor", which indicates a measure for general intelligence (Brand 1996), other research in the field of cognitive science explores intelligence in relation to the evolutionary experience of individuals. This means that, rather than having a general form of intelligence, humans become much more effective in solving problems that occur in the context of familiar situations (Wechsler 1964).

Another view on intelligence supposes that general human intelligence can be subdivided into specialized intelligence components, such as linguistic, logical-mathematical, musical, kinesthetic, spatial, social, or existential intelligence (Gardner 2000).

Synthesizing those perspectives on human intelligence, Sternberg (1985) proposes three distinctive dimensions of intelligence: componential, contextual, and experiential. The componential dimension of intelligence refers to some kind of individual (general) skill set of humans. Experiential intelligence refers to one´s ability to learn and adapt through evolutionary experience. Finally, contextual intelligence defines the capacity of the mind to inductively understand and act in specific situations as well as the ability to make choices and modify those contexts.

## 2.3 Collective Intelligence

Another related concept is collective intelligence. According to Malone and Bernstein (2015, p. 3), collective intelligence refers to "[…] groups of individuals acting collectively in ways that seem intelligent". Even though the term "individuals" leaves room for interpretation, researchers in this domain usually refer to the concept of wisdom of crowds and, thus, a combined intelligence of individual human agents (Woolley et al. 2010). This concept describes that, under certain conditions, a group of average people can outperform any individual of the group or even a single expert (Leimeister 2010). Other well-known examples of collective intelligence are phenomena found in biology, where, for example, a school of fish swerves to increase protection against predators (Berdahl et al. 2013). These examples show that collective intelligence typically refers to large groups of homogenous individuals (i.e., humans or animals), whereas Hybrid Intelligence combines the complementary intelligence of heterogeneous agents (i.e., humans and machines).

## 2.4 Artificial Intelligence

The subfield of intelligence that relates to machines is called artificial intelligence (AI). With this term, we mean systems that perform "[…] activities that we associate with human thinking, activities such as decision-making, problem solving, learning […]" (Bellman 1978, p. 3). It generally covers the idea of creating machines that can accomplish complex goals. The basic idea behind this concept is, that, by applying machine learning techniques, a system becomes capable of analyzing its environment and adapting to new circumstances in this environment. Examples for this are object recognition, problem solving, or natural language processing (Russell and Norvig 2016). Other streams of research in this domain perceive AI as the "[…] synthesis and analysis of computational agents that act intelligently […]" (Poole and Mackworth 2017, p. 3). Moreover, AI can be described as having the general goal to replicate the human mind by defining it as "[…] the art of creating machines that perform functions that require intelligence when performed by people […]" (Kurzweil 1990, p. 117). The performance of AI in achieving human-level intelligence can then be measured by, for instance, the Turing test. This test asks an AI program to simulate a human in a text-based conversation. However, due to the multi-facetted nature of general intelligence, such capabilities can be seen as a sufficient but not necessary criterion for artificial general intelligence (Searle 1980).

Synthesizing those various definitions in the field, AI includes elements such as the human-level ability to solve domain-independent problems, the capability to combine highly task-specialized and more generalized intelligence, the ability to learn from its environment and to interact with other intelligent systems, or human teachers, which allows intelligent agents to improve in problem solving through experience.

To create such a kind of AI in intelligent agents, various approaches exist that are more or less associated with the understanding and replication of intelligence. For instance, the field of cognitive computing "[…] aims to develop a coherent, unified, universal mechanism inspired by the mind's capabilities. […] We seek to implement a unified computational theory of the mind […] "(Modha et al. 2011, p. 60). Therefore, interdisciplinary research teams rely on the reverse-engineering of human learning to create machines that "[…] learn and think like people […]" (Lake et al. 2017, p. 1).

## 3 The Complementary Benefits of Human and Artificial Intelligence

The general rationale behind the idea of Hybrid Intelligence is that humans and computers have complementary capabilities that can be combined to augment each other. The tasks that can be easily done by artificial and human intelligence are quite divergent. This fact is known as Moravec´s paradox (1988, p. 15), which states that "[…] it is comparatively easy to make computers exhibit adult level performance on intelligence tests or playing checkers, and difficult or impossible to give them the skills of a one-year-old when it comes to perception and mobility […]". This is especially true for the human common sense that is challenging to achieve in AI (Lake et al. 2017).

This can be explained by the separation of two distinct types of cognitive procedures (Kahneman 2011). The first, system 1, is fast, automatic, affective, emotional, stereotypic, subconscious, and it capitalizes on what one might call human intuition. The second one, system 2, is rather effortful, logical, and conscious, and ideally follows strict rational rules of probability theory. In the context of complementary capabilities of human and artificial intelligence, humans have proved to be superior in various settings that require system 1 thinking. Humans are flexible, creative, empathic, and can adapt to various settings. This allows, for instance, human domain experts to deal with so called "broken-leg" predictions that deviate from the currently known probability distribution. However, they are restricted by bound rationality that prevents them from aggregating information perfectly and drawing conclusions from that. On the other hand, machines are particularly good at solving repetitive tasks that require fast processing of huge amounts of data, at recognizing complex patterns, or weighing multiple factors following consistent rules of probability theory. This has been proven by a long-standing tradition of research that shows the superiority of machines in such fields of application. Even in very simple actuarial models, they outperform human experts in making predictions under uncertainty (Meehl 1954). Figure 1 summarizes the two types of thinking as well as the respective strengths of humans and machines.

These complementary strengths of humans and machines (see Fig. 1) have since led to two different forms of interplay, that is, AI is in the loop of human intelligence, improving human decisions by providing predictions, and human intelligence is in the loop of AI, a form which is frequently applied to train machine learning models.

### 3.1 Artificial Intelligence in the Loop of Human Intelligence

Currently, in typical business contexts, AI is applied in two areas. First, it is used in automating tasks that can be solved by machines alone. While this is frequently associated with the fear of machines taking over jobs and making humans obsolete in the future, it might also allow machines to solve tasks that humans do not want to do themselves. Second, AI is applied to provide humans with decision support by offering some kind of prediction. This ranges from structuring data, making forecasts, for example, in financial markets, or even predicting the best set of hyperparameters to train new machine learning models (e.g., AutoML). As humans frequently act non-Bayesian by violating probabilistic rules and thus making inconsistent decisions, AI has proven to be a valuable tool to help humans in making better decisions (Agrawal et al. 2018). The goal in this context is to improve human decision effectiveness and efficiency.

In settings where AI provides us with input that is then evaluated to make a decision, humans and machines act as teammates. For instance, by processing patient data (e.g., CT scans) AI can help human physicians to make predictions on diseases such as cancer, thereby empowering the doctor to learn from the additional guidance. In this context, the Hybrid Intelligence approach allows human experts to leverage the predictive power of AI while using their own intuition and empathy to make a choice from the predictions of the AI.

### 3.2 Human Intelligence in the Loop of Artificial Intelligence

On the other hand, human intelligence also has a crucial role in the loop of machine learning and AI. In particular, humans provide assistance in several parts of the machine learning process to support AI in tasks that it cannot (yet) solve alone. Here, humans are most commonly needed for the generation of algorithms (e.g., hyperparameter setting/tuning), for training or debugging models, and for making sense of unsupervised approaches such as data clustering.

In this case, AI systems can benefit and learn from human input. This approach allows for integrating human domain knowledge in the AI to design, complement and evaluate the capabilities of AI (Mnih et al. 2015). Many of these applications are based on supervised and interactive learning approaches and require an enormous amount of labeled data, provided by humans (Amershi et al. 2014). The basic rationale behind this approach is that humans act as teachers who train an AI. The same machine teaching approach can also be found in the area of reinforcement learning that uses, for instance, human game play as input to initially train robots. In this context, human intelligence functions as a teacher, augmenting the AI. Hybrid Intelligence allows to distribute computational tasks to human intelligence on demand (e.g., through crowdsourcing) to minimize shortcomings of current AI systems. Such human-in-the-loop approaches are particularly valuable when only little data is available. In addition, they can be used when pre-trained models need to be adapted for specific domains, or in contexts where human annotations are already used.

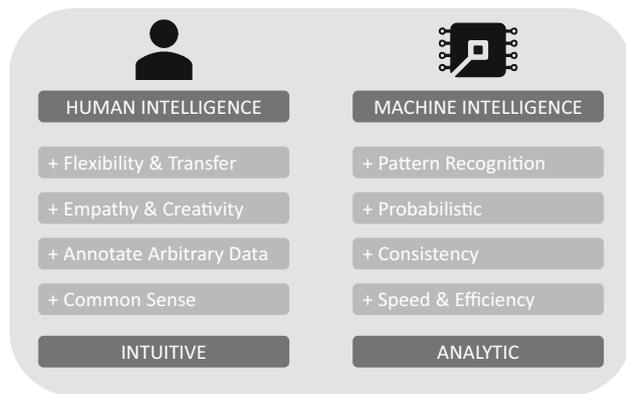

Fig. 1 Complementary strengths of humans and machines

Since human intelligence in the loop of AI is most frequently applied in settings where models are initially set up or in the field of research, the goal is to make AI more effective. Figure 2 summarizes the distribution of roles in Hybrid Intelligence.

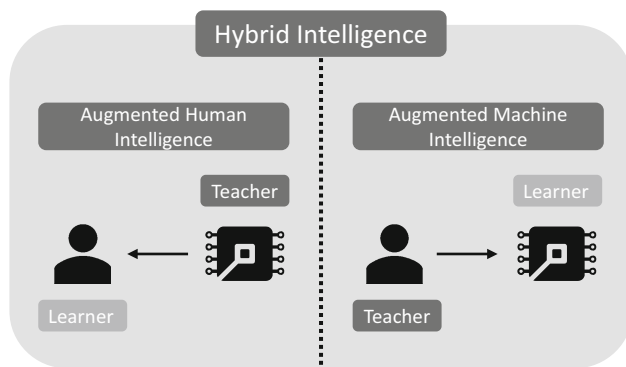

Fig. 2 Distribution of roles in hybrid intelligence

## 4 Defining Hybrid Intelligence

Another approach is to combine human and artificial intelligence. The basic rationale behind this is the combination of complementary heterogeneous intelligences (i.e., human and artificial agents) to create a socio-technological ensemble that is able to overcome the current limitations of (artificial) intelligence. This approach focuses neither on human intelligence in the loop of AI nor on automating simple tasks through machine learning. Rather, the emphasis lies on solving complex problems using the deliberate allocation of tasks among different heterogeneous algorithmic and human agents. Both the human and the artificial agents of such systems can then co-evolve by learning and achieve a superior outcome on the system level.

In accordance with Dellermann et al. (2019), we call this concept Hybrid Intelligence, which is defined as the ability to achieve complex goals by combining human and artificial intelligence, thereby reaching superior results to those each of them could have accomplished separately, and continuously improve by learning from each other.[1] Several core concepts of this definition are noteworthy:

- *Collectively* Hybrid Intelligence covers the fact that tasks are performed collectively. Consequently, activities conducted by each agent are conditionally dependent. However, their goals are not necessarily always aligned to achieve the common goal such as when humans are teaching an AI adversarial tactics in playing games.
- *Superior results* This defines the idiosyncratic fact that the socio-technical system achieves a performance in a specific task that none of the involved agents, whether they are human or artificial, could have achieved without the other. The aim is, therefore, to make the outcome (e.g., a prediction) both more efficient and effective on the level of the socio-technical system by achieving goals that could not have been solved before. This contrasts Hybrid Intelligence with the most common applications of human-in-the-loop machine learning.
- *Continuous learning* a central aspect of Hybrid Intelligence is that, over time, this socio-technological system improves, both as a whole and each single component (i.e., human and artificial agents). This facet shows that they learn from each other through experience. The performance of Hybrid Intelligence systems can, thus, not only be measured by the superior outcome of the whole socio-technical system alone, but the learning (i.e., performance increase) of human

---

[1] For further work on this topic see Dellermann et al. (2019).

and machine agents that are parts of the system must also be taken into account.

Figure 3 displays the conceptual integration of Hybrid Intelligence in related fields of research and the concepts discussed earlier in the paper.

One recent example that provides an astonishing indicator for the potential of Hybrid Intelligence is DeepMind[2]´s AlphaGo. For training the game-playing AI, a supervised learning approach was used that learned from expert human moves and, thus, augment and improve the AI through human input, which allowed AlphaGo to achieve superhuman performance over time. During its games against various human world-class players, AlphaGo played several highly creative moves that previously were beyond human players´ imagination. Consequently, AlphaGo was able to augment human intelligence as well and somehow taught expert players completely new knowledge in a game that is one of the longest studied in human history (Silver et al. 2016).

> I believe players more or less have all been affected by Professor Alpha. AlphaGo's play makes us feel more free and no move is impossible to play anymore. Now everyone is trying to play in a style that hasn't been tried before. – **Zhou Ruiyang, 9 Dan Professional, World Champion**

Solving problems through Hybrid Intelligence offers the possibility to allocate a task between humans and artificial agents, and deliberately achieve a superior outcome on the socio-technical system level by aggregating the output of its parts. Moreover, such systems can improve over time by learning from each other through various mechanisms, such as labeling, demonstrating, teaching adversarial moves, criticizing, rewarding and so on. This will allow us to augment both the human mind and the AI and extend applications when men and machines can learn from each other in much more complex tasks than games: for instance, strategic decision making, managerial, political or military decisions, science, and even AI development leading to AI reproducing itself in the future. Hybrid Intelligence, therefore, offers the opportunity to achieve super-human levels of performance in tasks that so far seem to be at the core of human intellect.

## 5 The Advantages of Hybrid Intelligence

This hybrid approach provides various advantages for humans in the era of AI such as generating new knowledge in complex domains that allow humans to learn from AI and transfer implicit knowledge from experienced experts to novices without any kind of social interaction. On the other hand, the human teaching approach makes it possible to control the learning process by ensuring that the AI makes inferences based on criteria that can be interpreted by humans – a fact that is crucial for AI adoption in many real-world applications and AI safety and that makes it possible to exclude biases such as racism (Bostrom 2017). Moreover, such hybrid approaches might allow for a better customization of AI, based on learning the preferences of humans during interaction. Finally, we argue that the co-creation of Hybrid Intelligence services between humans and intelligent agents might create a sense of psychological ownership and, thus, increase acceptance and trust.

## 6 Future Research Directions for the BISE Community

As the technological development continues, the focus of machine learning and Hybrid Intelligence is shifting towards applications in real-world business contexts, but solving complex problems will become the next challenge. Such complex problems in managerial settings are typically time variant, dynamic, require much domain knowledge and have no specific ground truth. These highly uncertain contexts require intuitive and analytic abilities, as well as human strengths such as creativity and empathy. Consequently, we propose three specific but also interrelated directions for further development of the concept in the field of BISE that are focused on socio-technical system design.

First, a lack of trust in AI is one of the most challenging barriers to AI adoption. Furthermore, we need to keep in mind that we should not aim at maximizing trust in AI, but rather find a balance between trust and distrust that makes it possible to leverage the potentials of AI and at the same time avoids negative effects stemming from overreliance on AI (Lee and See 2004). We believe this challenge can be overcome by researchers in the field of Hybrid Intelligence, since a key requirement for integrating human input into an AI system is the translation of a system's state and needs in a way that humans can understand and process them accordingly and vice versa. For instance, semi-autonomous driving requires the AI to sense the state of the human in order to distribute tasks between itself and the human driver. Furthermore, it requires examining human-centered AI architectures that balance, for instance, transparency of the underlying model and its performance. However, domain specific design guidelines for developing user-interfaces that allow humans to understand and process the needs of an artificial system are still missing. We, therefore, believe that more research is needed to develop more suitable human-AI interfaces as well as to investigate possible task and interface designs that allow human helpers to teach an AI system (e.g., Simard et al. 2017). Ensuring interpretability and transparency of machine learning models while maintaining accuracy is one of the most crucial challenges in research on Hybrid Intelligence, since it is one key foundation for building appropriate trust in AI. This was most recently covered by the launch of the People + AI Research (PAIR[3]) group at Google brain, which indicates the high relevance for both academia and practice.

---

[2] https://deepmind.com (accessed 19 Mar 2019).

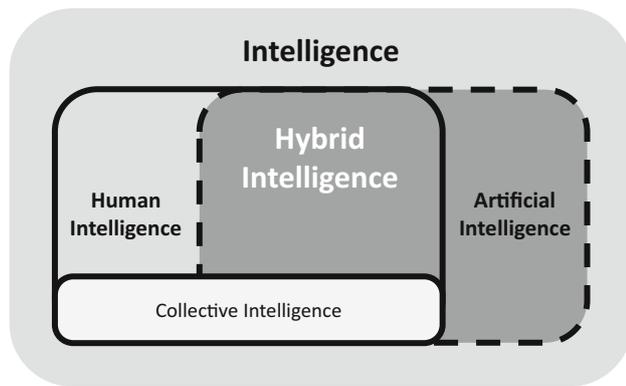

**Fig. 3** Conceptual integration of hybrid intelligence

Second, research in the field of Hybrid Intelligence might investigate what kind of governance mechanisms can be used to train and maintain Hybrid Intelligence systems. Such tasks frequently require domain expertise (e.g., health care) and, thus, system designers need to focus on explicitly matching experts with tasks, aggregating their input and assuring quality standards. We, therefore, argue that it might be a fruitful area of research to further investigate which kind of governance mechanisms might be applicable in Hybrid Intelligence systems. Moreover, human teachers may have different motivations to contribute to the system. Consequently, research in the field tries to shed light on the question of how to design the best incentive structure for a predefined task. Especially, when highly educated and skilled experts are required to augment AI systems, the question arises if traditional incentives of micro-tasking platforms (e.g., monetary reward) or online communities (e.g., social rewards) are sufficient.

A third avenue for future research is related to digital work mechanisms. The rise of AI is now changing the capabilities of IS and the potential distribution of tasks between human and IS dramatically and, hence, affects the core of our discipline. Those changes create novel qualification demands and skill sets for employees and, consequently, provide promising directions for IS education. Such research might examine the educational requirements for democratizing the use of AI in future workspaces.

Finally, Hybrid Intelligence also offers great possibilities for novel forms of digital work such as internal crowd work to leverage the collective knowledge of individual experts that resides within a company across functional silos.

## References

Agrawal A, Gans J, Goldfarb A (2018) Prediction machines: the simple economics of artificial intelligence. Harvard Business Press, Boston

Amershi S, Cakmak M, Knox WB, Kulesza T (2014) Power to the people: the role of humans in interactive machine learning. AI Mag 35(4):105–120

Bellman R (1978) An introduction to artificial intelligence: can computers think?. Boyd & Fraser, San Francisco

Berdahl A, Torney CJ, Ioannou CC, Faria JJ, Couzin ID (2013) Emergent sensing of complex environments by mobile animal groups. Science 339(6119):574–576

Bostrom N (2017) Superintelligence. Dunod, Paris

Brand C (1996) The g factor: general intelligence and its implications. Wiley, Hoboken

Dellermann D, Calma A, Lipusch N, Weber T, Weigel S, Ebel P (2019) The future of human-ai collaboration: a taxonomy of design knowledge for hybrid intelligence systems. In: Hawaii international conference on system sciences (HICSS). Hawaii, USA

Gardner HE (2000) Intelligence reframed: multiple intelligences for the 21st century. Hachette, London

Goodfellow I, Bengio Y, Courville A (2016) Deep learning. MIT Press, Cambridge

Gottfredson LS (1997) Mainstream science on intelligence: an editorial with 52 signatories, history, and bibliography. Intelligence 24(1):13–23

Kahneman D (2011) Thinking, fast and slow. Macmillan, London

Kamar E (2016) Hybrid workplaces of the future. XRDS 23(2):22–25

Kurzweil R (1990) The age of intelligent machines. MIT Press, Cambridge

Lee JD, See KA (2004) Trust in automation: designing for appropriate reliance. Hum Factor 46(1):50–80

Leimeister JM (2010) Collective intelligence. Bus Inf Syst Eng 2(4):245–248

Malone TW, Bernstein MS (2015) Handbook of collective intelligence. MIT Press, Cambridge

McAfee A, Brynjolfsson E (2017) Machine, platform, crowd: harnessing our digital future. WW Norton & Company, New York

Medsker LR (2012) Hybrid intelligent systems. Springer, Heidelberg

Meehl PE (1954) Clinical versus statistical prediction: a theoretical analysis and a review of the evidence. University of Minnesota Press, Minneapolis

Mnih V, Kavukcuoglu K, Silver D, Rusu AA, Veness J, Bellemare MG, Graves A, Riedmiller M, Fidjeland AK, Ostrovski G, Petersen S, Beattie C, Sadik A, Antonoglou I, King H, Kumaran D, Wierstra D, Legg S, Hassabis D (2015) Human-level control through deep reinforcement learning. Nature 518(7540):529–533

Modha DS, Ananthanarayanan R, Esser SK, Ndirango A, Sherbondy AJ, Singh R (2011) Cognitive computing. Commun ACM 54(8):62–71

Moravec H (1988) Mind children: The future of robot and human intelligence. Harvard University Press, Cambridge

---

[3] https://ai.google/research/teams/brain/pair (accessed 19 Mar 2019).

Poole DL, Mackworth AK (2017) Artificial intelligence: foundations of computational agents, 2nd edn. Oxford University Press, Oxford

Russell SJ, Norvig P (2016) Artificial intelligence: a modern approach. Pearson Education Limited, London

Searle JR (1980) Minds, brains, and programs. Behav Brain Sci 3(3):417–424

Silver D, Huang A, Maddison CJ, Guez A, Sifre L, van den Driessche G, Schrittwieser J, Antonoglou I, Panneershelvam V, Lanctot M (2016) Mastering the game of Go with deep neural networks and tree search. Nature 529(7587):484–489

Simard PY, Amershi S, Chickering DM, Pelton AE, Ghorashi S, Meek C, Ramos G, Suh J, Verwey J, Wang M, Wernsing J (2017) Machine teaching: a new paradigm for building machine learning systems. CoRR abs/1707.06742

Sternberg RJ (1985) Beyond IQ: a triarchic theory of human intelligence. Cambridge University Press, Cambridge, England

Ullman T, Tenenbaum J, Gershman SJ (2017) Building machines that learn and think like people. Behav Brain Sci. https://doi.org/10.1017/S0140525X16001837

Wechsler D (1964) Die Messung der Intelligenz Erwachsener. Huber, Bern

Woolley AW, Chabris CF, Pentland A, Hashmi N, Malone TW (2010) Evidence for a collective intelligence factor in the performance of human groups. Science 330(6004):686–688